\documentclass[journal,twoside,web]{ieeecolor}
\usepackage{generic}
\usepackage{cite}
\usepackage{amsmath,amssymb,amsfonts}
\usepackage{algorithmic}
\usepackage{graphicx}
\usepackage{textcomp}

\usepackage{amsmath,amssymb,amsfonts}
\usepackage{graphicx}
\usepackage{textcomp}
\usepackage{soul}
\usepackage[colorlinks,allcolors=black,pdftex]{hyperref}

\usepackage{booktabs} 

\def\BibTeX{{\rm B\kern-.05em{\sc i\kern-.025em b}\kern-.08em
    T\kern-.1667em\lower.7ex\hbox{E}\kern-.125emX}}
\begin{document}
\title{Adversarial Vessel-Unveiling Semi-Supervised Segmentation for Retinopathy of Prematurity Diagnosis}

\author{Gozde Merve Demirci, Jiachen Yao, Ming-Chih Ho, Xiaoling Hu, Wei-Chi Wu, Chao Chen, and Chia-Ling Tsai
\thanks{G. Demirci with the Computer Science Department, Graduate Center, CUNY, New York, NY 11367 USA, (e-mail: gdemirci1@gradcenter.cuny.edu).}
\thanks{J. Yao, with the Computer Science Department, Stony Brook University, SUNY, Stony Brook, NY 11794-2424 USA (email: jiacyao@cs.stonybrook.edu)}
\thanks{C. Chen, with the Department of Biomedical Informatics, Stony Brook University, SUNY, Stony Brook, NY 11794-2424 USA (email: chao.chen.1@stonybrook.edu)}
\thanks{X. Hu, with the Harvard Medical School, Harvard University, Boston, MA 02115 USA (email: xihu3@mgh.harvard.edu)}
\thanks{M.-C. Ho and W.-C. Wi Chang Gung Memorial Hospital, Taoyuan, Taiwan (e-mail: weichi666@gmail.com; d4352507@yahoo.com.tw; ).}
\thanks{C.-L. Tsai with the Computer Science Department, Queens College, CUNY, Queens, NY 11367 USA, (e-mail: chialing.tsai@qc.cuny.edu).}}
\maketitle

\begin{abstract}
Accurate segmentation of retinal images plays a crucial role in aiding ophthalmologists in diagnosing retinopathy of prematurity (ROP) and assessing its severity. However, due to their underdeveloped, thinner vessels, manual annotation in infant fundus images is very complex, and this presents challenges for fully-supervised learning. To address the scarcity of annotations, we propose a semi-supervised segmentation framework designed to advance ROP studies without the need for extensive manual vessel annotation. Unlike previous methods that rely solely on limited labeled data, our approach leverages teacher-student learning by integrating two powerful components: an uncertainty-weighted vessel-unveiling module and domain adversarial learning. The vessel-unveiling module helps the model effectively reveal obscured and hard-to-detect vessel structures, while adversarial training aligns feature representations across different domains, ensuring robust and generalizable vessel segmentations.
We validate our approach on public datasets (CHASEDB, STARE) and an in-house ROP dataset, demonstrating its superior performance across multiple evaluation metrics. Additionally, we extend the model's utility to a downstream task of ROP multi-stage classification, where vessel masks extracted by our segmentation model improve diagnostic accuracy. The promising results in classification underscore the model’s potential for clinical application, particularly in early-stage ROP diagnosis and intervention. Overall, our work offers a scalable solution for leveraging unlabeled data in pediatric ophthalmology, opening new avenues for biomarker discovery and clinical research.
\end{abstract}


\section{Introduction}
\label{sec:introduction}

Retinopathy of prematurity (ROP) is one of the leading causes of vision impairment and blindness in preterm infants. It occurs when abnormal blood vessels develop in the retina, which is the layer of tissue at the back of the eye that is sensitive to light~\cite{rop_intro}. Fundus images are often used to diagnose the presence and severity of ROP since they provide information about the structure and function of capillaries~\cite{b63, b64}. Being a vascular disease, ROP diagnosis heavily depends on vessel morphology and distribution, in particular, vessel tortuosity, vessel coverage, and the progression of neovascularization, which refer to the twisting and turning of blood vessels, the extent to which blood vessels cover the retina, and the growth of abnormal blood vessels, respectively. Therefore, vessel extraction is a crucial step for ROP study.

Manual annotation of vessels from ROP fundus images is extremely challenging; images of preterm babies are characterized by their underdeveloped retinal vessels and choroidal vessels underneath retinal pigment epithelium. Previous studies trained automatic vessel segmentation models for ROP with proprietary datasets~\cite{pete_vessel,b63,b75} with high-quality annotation, but such models are not available to the public. Therefore, leveraging publicly available labeled data remains essential for effectively segmenting unlabeled ROP data.

\begin{figure}[!hbt]
\centerline{\includegraphics[width=\columnwidth]{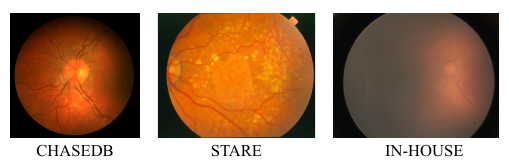}}
\caption{The domain difference among publicly available datasets (CHASEDB \cite{b27} \& STARE \cite{b28}) and our ROP dataset. CHASEDB primarily consists of images from adolescents and healthy populations, while STARE comprises images from adults and emphasizes heterogeneous pathology cases. In contrast, our ROP dataset is curated for infants' retinopathy of prematurity detection.}
\label{fig:dataset}
\end{figure}

We introduce a semi-supervised learning (SSL) approach to address the ROP vessel segmentation problem, effectively leveraging both publicly labeled datasets and unlabeled ROP data.
SSL has shown promises in various medical imaging applications~\cite{b32, b44, b45}, including retinal vessel segmentation~\cite{b41, b42}. Compared to other SSL techniques~\cite{dual_teacher, b60, b58, b51}, we used a teacher-student network, where the student learns more aggressively from both labeled and unlabeled data to explore new features, while the teacher focuses solely on the unlabeled target data, evolving conservatively through an Exponential Moving Average (EMA) of the student’s weights. This targeted knowledge transfer enhances ROP vessel extraction.


Despite advancements in teacher-student networks, segmenting ROP images remains challenging due to both external and internal factors. The \textbf{\textit{external challenge}} stems from the significant distribution gap between public datasets and ROP images. Public datasets~\cite{b27,b28} differ significantly from ROP images in terms of age population and disease diversity. These datasets mostly consist of images from adolescents and adults with prevalent conditions like diabetic retinopathy and glaucoma~\cite{b27, b28, b70, b71, b72, b73, b74}, where the vessels are thicker and have better contrast with the background. In contrast, ROP images from preterm infants have underdeveloped retinas, and exhibit a more diffuse, white-ish tissue appearance with thinner, less developed vessels, as shown in Figure~\ref{fig:dataset}. The \textbf{\textit{internal challenge}} arises from the high variability in ROP imaging conditions. While vessels near the optic disc (OD) are clearer, image quality deteriorates further from the OD, and there is substantial variability in ROP images due to the diverse developmental stages of preterm infants.

Considering these challenges and the lack of ROP image annotations, applying the conventional SSL approach would struggle to exploit the available labeled data from public datasets fully. The unique characteristics of ROP images, such as the underdeveloped retinal structures, already complicate the segmentation process. Unsupervised methods, like domain adaptation, may also overlook valuable segmentation cues from labeled public data, which are crucial in guiding the model without ROP labels. Therefore, to mitigate the external challenge, we incorporated a domain adversarial framework into our semi-supervised learning. Domain adversarial techniques have demonstrated promising improvements [37], [38], including in medical imaging [39], and particularly in retinal images [40], [41]. We use an adversarial discriminator to align the latent feature representations from both domains, ensuring they remain informative for vessel segmentation while reducing the domain gap by aligning feature distributions between the two domains through adversarial training. This feature alignment helps reduce irrelevant attributes such as color and background variations, improving the vessel extraction performance on the unlabeled ROP domain. This allows us to balance domain generalization and specificity, extracting meaningful vessel patterns across both domains. For the internal challenge, we introduce an uncertainty-weighted vessel-unveiling module that applies an uncertainty map to enhance hidden or obscured retinal vessels in ROP images, particularly in areas with variable imaging conditions. This module maximizes vessel coverage, which is critical for accurate ROP diagnosis.



This paper proposes an adversarial feature alignment-based uncertainty-aware vessel-unveiling SSL method. 
 To the best of our knowledge, this is the first attempt to combine adversarial feature alignment in SSL with a vessel-uncertainty map to enhance vessel visibility and ensure comprehensive vessel extraction for ROP segmentation.

In our work, we make four three contributions: \begin{itemize} 

\item We propose the first to apply semi-supervised learning to ROP vessel segmentation, utilizing publicly available labeled data alongside unlabeled ROP images to enhance segmentation accuracy.

\item We propose a teacher-student network combined with an adversarial feature alignment discriminator to address the domain gap between public datasets and ROP images for vessel segmentation. 

\item We introduce an uncertainty-aware vessel-unveiling module that enhances underdeveloped and obscured vessels, improving segmentation performance in ROP images. 

\item This is the first study to segment all four stages of ROP images without ground-truth vessel annotations and to use the segmented masks in a downstream task, emphasizing adaptability to real-world clinical tasks. 

\end{itemize}

\section{Related Works}

\subsection{Semi-Supervised Medical Segmentation}
Semi-supervised learning strategies encompass a range of techniques, broadly falling into three categories: self-training~\cite{b50,b51}, consistency-based learning~\cite{b22,b11,b58}, and generative models~\cite{b7,b10,b52,b53}. 

In the self-training paradigm, models initially train on a limited labeled dataset and subsequently leverage predictions on a larger, unlabeled dataset to enhance accuracy~\cite{b50}. A notable enhancement to this approach is introduced in~\cite{b51}, where the authors implement a robust self-training model by incorporating strong data augmentations (SDA) on unlabeled images. Consistency-based learning methods, exemplified by Mean Teacher (MT)~\cite{b22}, enforce coherence between predictions on augmented and original data. MT adopts a teacher-student architecture, integrating perturbation-based consistency loss on unlabeled data alongside supervised loss on labeled data, leading to the generation of consistent pseudo-labels. 

Generative models, including generative adversarial networks (GAN) and variational autoencoders, focus on synthesizing training data by modeling the input data distribution~\cite{b52,b53}. GAN, with its discriminator and generator components, refines the segmentation network output iteratively until it becomes indistinguishable from the ground truth. Adversarial learning has gained prominence as an effective strategy for leveraging unlabeled data in semi-supervised medical image segmentation. A significant contribution is seen in~\cite{b10}, featuring a framework with two generators and a discriminator. The discriminator evaluates masks from generators, extracting supervised information that aids learning from unannotated data. 

\subsection{Adversarial Domain Frameworks}

Adversarial domain frameworks have emerged as a powerful approach to address domain shifts in various tasks, including medical imaging, where labeled data is scarce or expensive to annotate in the target domain. These techniques aim to reduce the domain gap between a source domain and a target domain by aligning distributions in different levels such as feature, input, and output, while preserving the task-relevant semantics.~\cite{dann_ganin}'s seminal work on Domain-Adversarial Neural Networks (DANN) introduced a gradient reversal layer to achieve domain-invariant feature representations, which laid the foundation for subsequent work in this area. Following this, several domain adaptation techniques have been successfully applied to medical imaging tasks~\cite{uda_wei, da_survey}, showing significant improvements in cross-domain performance, particularly when addressing domain shifts between different modalities or datasets.

In medical imaging, domain adversarial approaches effectively mitigate domain gaps caused by varying imaging devices, patient populations, or disease characteristics. For instance,~\cite{cross_xiao} and ~\cite{cross_weng} applied adversarial frameworks to align features between source datasets of healthy retinal images and target datasets with pathological conditions, improving segmentation and classification tasks. Additionally, ~\cite{cross_Lv} integrated the teacher-student network with adversarial frameworks to improve segmentation in cross-domain scenarios, such as brain MRI segmentation. Despite these advancements, segmenting ROP vessels introduces unique challenges due to the inherent differences in retinal vessel morphology and imaging quality between public datasets and ROP fundus images.

\subsection{ROP Vessel Segmentation}

Segmenting preterm fundus images is crucial for diagnosing ROP, especially in severe stages where vessel tortuosity and structure are significant indicators. However, this task is challenging due to the underdeveloped morphological structures of preterm infants. Several techniques have been applied to extract vessel structure through preterm fundus images for ROP diagnosis. \cite{b75} applied image-processing steps and thresholding to skeletonize vessels for ROP plus disease, an ROP condition focusing on vessel abnormalities. These skeletonized vessels are then manually arranged to detect the presence of plus disease in retinal images. Additionally, deep learning methods have been a great help in extracting vessel segmentation for ROP detection. For instance, \cite{rop_seg} and \cite{b63} used fully supervised segmentation architectures to create vessel segmentation from private datasets with limited manually annotated ground truths. Despite their good performance, these methods could not fully realize the potential of deep learning models due to the small dataset size.

Within the context of small annotated datasets, one could utilize semi-supervised segmentation models to leverage a small amount of labeled data alongside a large amount of unlabeled data.  Although one publicly available ROP fundus image dataset exists~\cite{rop_data}, the vessel annotations, though valuable, tend to be structurally imprecise, sometimes missing major vessels or drawing others thicker than their true size. As a result, this dataset was not suitable as an annotated set for semi-supervised segmentation in our study. To the best of our knowledge, no previous work in the literature has applied semi-supervised segmentation to address ROP vessel segmentation without labeled ROP images.

\section{Method}

\begin{figure*}[!hbt]
\center
\centerline{\includegraphics[width=0.82\textwidth]{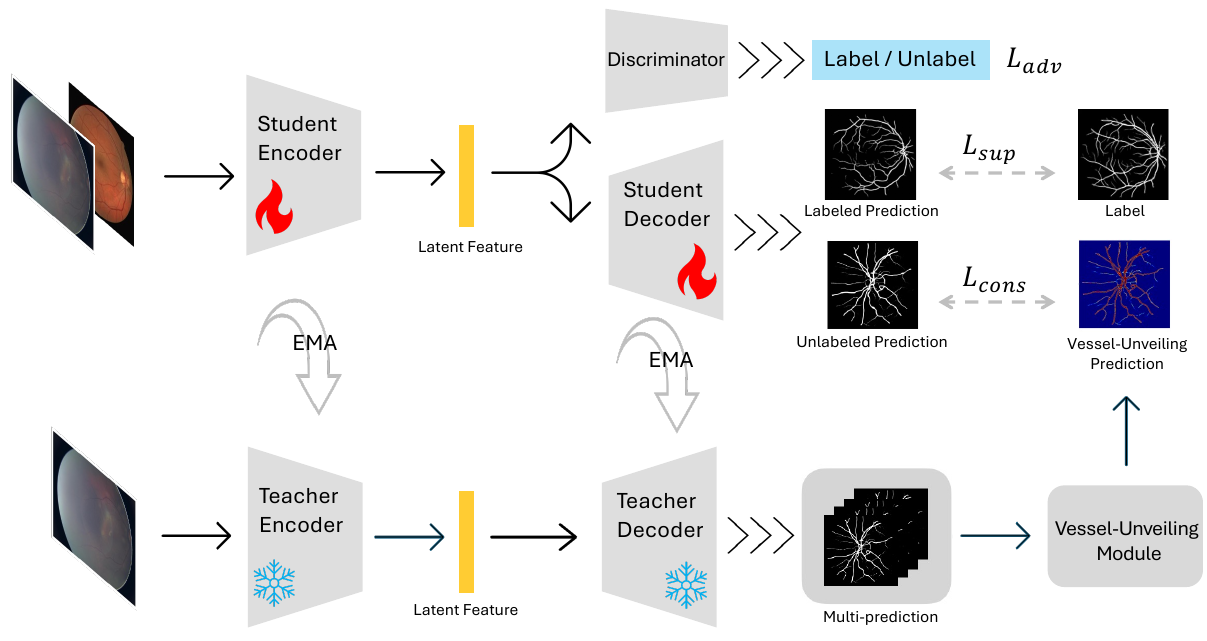}}
\caption{The framework of our proposed semi-supervised learning method for vessel segmentation. The student network learns from the supervised and consistency loss. The vessel-unveiling prediction extracted from the uncertainty-aware vessel-unveiling module reveals the hidden vessels to ensure boosted student network's performance on unlabeled inputs. The discriminator takes latent representation of both labeled and unlabeled domain to close the domain gap by aligning the features.}
\label{fig1}
\end{figure*}

The task of semi-supervised vessel segmentation is defined as follows: The training process involves utilizing dataset $D$, which consists of both labeled and unlabeled samples. Specifically, the dataset $D$ comprises a small set of $N$ labeled retinal samples denoted as $D^L = \{x^i, y^i\}_{i=1}^N$ and a substantially larger set of $M$ unlabeled retinal image samples, denoted as $D^U = \{x^j\}_{j=1}^M$. In this context, $x^i$ and $x^j$ represent fundus images, while $y^i$ denotes the corresponding ground truth segmentation mask. The objective is to utilize both $D^L$ and $D^U$ to improve segmentation performance across domains. Figure~\ref{fig1} overviews our semi-supervised segmentation framework, which comprises student and teacher models, as well as a discriminator to manage the domain gap. Both the teacher and student models adopt the U-Net architecture. The student model is trained with labeled and unlabeled data, while the teacher model is an EMA of the student model’s weights. The discriminator network is inspired by PatchGAN~\cite{b24}, consisting of convolutional layers designed to assess segmentation results for adversarial learning.

In our study, the optimization of the student model involves minimizing a loss function that consists of three distinct components, defined in Eq~\ref{student}.
\begin{equation}
L = L_{sup} + \alpha * L_{cons} + L_{adv}
\label{student}
\end{equation}
Firstly, the supervised segmentation loss ($L_{sup}$) is computed on labeled data ($D^L$), capturing the discrepancy between predicted segmentation and ground truth. Secondly, the consistency loss ($L_{cons}$) is introduced, calculated from the teacher model's output on unlabeled data ($D^U$), aiming to capture hidden vessels among the teacher's predictions and guide the student model and aligns the student and teacher model's outputs calculated on $D^U$. Additionally, we incorporate a third loss component, the adversarial loss ($L_{adv}$) derived from the discriminator branch to align the features in feature embedding. 

\subsection{Supervised Segmentation Loss}

Given the limited number of labeled image pairs, developing a robust model poses challenges of overfitting and lack of variability. To address these, we modify the student network’s architecture by incorporating three distinct decoders: one that retains the original U-Net decoder structure and two regularization decoders that apply feature noise and feature dropout.

The feature noise decoder introduces multiplicative noise to the feature maps. For each latent feature $\mathbf{z} \in \mathbb{R}^{H \times W \times C}$, we sample a uniform noise $\varepsilon\sim \mathcal{U}^d(-\sigma, \sigma)$, where $\sigma$ is the scale. The noisy feature map is computed as $\mathbf{z}_{\text{noise}} = \mathbf{z} + \varepsilon$. This operation encourages the network to learn more robust representations by adding controlled noise to intermediate feature layers, making the model less sensitive to small variations.

The feature dropout decoder drops regions of the latent feature. The dropout is based on a dynamically computed threshold over the attention map, which cancels the influence of low attention area. First, the attention map \( A(\mathbf{z}) \) is calculated as the mean across latent feature channels; \[A(\mathbf{z}) = \frac{1}{C} \sum_{c=1}^{C} \mathbf{z}_c\] where \( C \) is the number of channels. 
A dropout mask \( M(\mathbf{z}) \) is generated based on a threshold $t$ over the attention map $M(\mathbf{z}) = \mathbb{I}(A(\mathbf{z}) < t)$. The feature map is updated by multiplying with the mask $\mathbf{z}_{\text{drop}} = M\cdot \mathbf{z} $. This method forces the model to rely on multiple features rather than overfitting to specific dominant regions.

Together, these two regularization decoders enhance generalization by reducing overfitting~\cite{b16} and improving performance on particularly crucial given the small labeled sample size.

The supervised loss for the student network is computed using labeled data \( D^L \), where each labeled sample consists of a fundus image \( x^i \) and ground truth segmentation mask \( y^i \). The student network generates three predicted masks \( \hat{y}_m^i \), \( \hat{y}_n^i \), and \( \hat{y}_d^i \) from the main decoder, noise decoder, and dropout decoder, respectively. The supervised loss \( L_{sup} \) is then calculated by averaging the binary cross-entropy and Dice loss across the outputs of all three decoders shown in Eq~\ref{sup}:

\begin{equation}
L_{sup} = \frac{1}{3} \sum_{k \in \{m, n, d\}} \left( \text{BCE}(\hat{y}_k^i, y^i) + \text{DICE}(\hat{y}_k^i, y^i) \right)
\label{sup}
\end{equation}

$BCE(.)$ and $DICE(.)$ represents binary-cross entropy loss and dice loss, respectively. This loss function measures the discrepancy between the predicted and ground truth segmentation masks. The extracted segmentation mask is subsequently fed into the discriminator to compute the adversarial loss, as elaborated in the forthcoming subsection.

\subsection{Vessel Unveiling Consistency}
In the task of generating segmentation outputs from the unlabeled ROP images ($D^U$), the teacher network faces the challenge of producing reliable predictions due to the absence of annotations, which often leads to noisy and uncertain outputs~\cite{b18}. To address this, we introduce an uncertainty-aware vessel-unveiling module within the teacher network during training, aimed at enhancing vessel extraction from ROP images. This scheme involves generating multiple predictions from the teacher model for the same input image under diverse perturbations and soft augmentations and creates vessel-unveiling prediction. The vessel-unveiling consistency loss $L_{cons}$ takes outputs from this module and the student network outputs.

To facilitate this process, we incorporate the MC Dropout technique~\cite{b9} into the teacher network for improved information extraction. Specifically, we perform K stochastic forward passes through the teacher model, applying random dropout and soft augmentation in each pass, resulting in a set of prediction probabilities ${p_k}$ for the same input. From these probabilities, we compute \textit{vessel-entropy} by estimating entropy values for ${p_k}$, which helps quantify uncertainty and extract as much vessel information as possible for each pixel.

\begin{figure}[!hbt]
\center
\centerline{\includegraphics[width=0.4\textwidth]{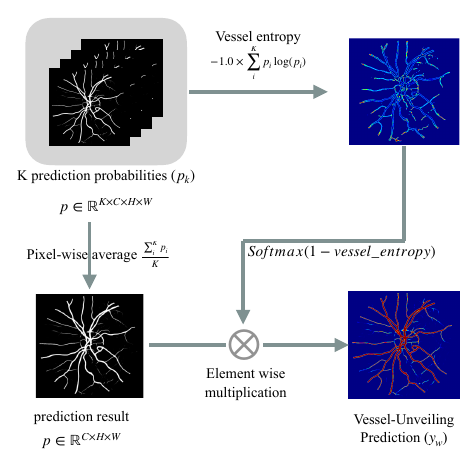}}
\caption{Uncertainty-aware vessel-unveiling Module. The vessel-unveiling reveals hidden vessels due to ROP image domain internal challenges discussed in section~\ref{sec:introduction}. In the heatmap scale, the red color represents a higher value.}
\label{uncert_fig}
\end{figure}

The entropy values are computed at the pixel level from the ROP images, resulting in a map of shape $\mathbb{R}^{H \times W \times 1}$, where $H$ and $W$ correspond to the image height and width, respectively. The softmax of the extracted vessel-entropy is then element-wise multiplied with the main segmentation prediction (obtained by averaging ${p_k}$) for the same input image. This operation serves to emphasize vessel regions with higher uncertainty in the teacher model's prediction on the unlabeled data, thus refining the vessel segmentation output. The resulting vessel-unveiling prediction from the teacher model for the unlabeled input is denoted as $y_w$. Figure~\ref{uncert_fig} illustrates the vessel-entropy extraction process and the final $y_w$ result.

The final consistency loss is defined as:
\begin{equation}
        L_{cons} = \alpha \times MSE(\hat{y}_j, \hat{y}'_j) + DIST(\hat{y}_j, y_w)
\label{total_consistency}
\end{equation}

where $\alpha$ is a weighting factor used to balance the two components of the loss. In this equation, the student model's output from the main decoder is represented as ${\hat{y}_j}$, the teacher model's output as ${\hat{y}'_j}$, and the uncertainty-aware vessel-unveiling prediction from the teacher model is denoted as $y_w$, all for the same unlabeled image. Each loss is formally defined in Equation~\ref{similarity} and Equation~\ref{uncertainty}, respectively. The vessel entropy, $I{vessel}$, which is referenced in Figure~\ref{uncert_fig}, plays a role in calculating this process.

\begin{equation}
MSE(\hat{y}_j, \hat{y}'_j) = \frac{1}{H \times W}\sum(\hat{y}_j - y_w)^2
\label{similarity}
\end{equation}

\begin{equation}
        DIST(\hat{y}_j, y_w) = \frac{1}{H \times W}\sum((\hat{y}_j - y_w)^2) \times (1 - I_{vessel})
\label{uncertainty}
\end{equation}

\subsection{Feature-Level Adversarial Domain Alignment}

Inspired by the PatchGAN approach~\cite{b24}, our discriminator produces a matrix output to capture fine-grained details and identify inconsistencies across domains. The discriminator's primary goal is to distinguish between labeled and unlabeled datasets, thus helping align the domain features for better generalization. To achieve this, we leverage the feature representations extracted from both labeled and unlabeled images via the student and teacher models' encoders, respectively. Despite differences in domain populations, the common modality ensures that the feature space preserves similar morphological characteristics, especially for vessel segmentation. This approach allows the discriminator to evaluate feature similarities between the public dataset and the ROP dataset, capturing intricate details of fundus images and improving segmentation accuracy. In this adversarial learning process, two inputs are involved: (1) a feature representation extracted from the student model of the labeled input image and (2) a feature representation extracted from the teacher model of the unlabeled input image.

The adversarial mechanism refines the student model's segmentation by providing critical feedback from the discriminator via adversarial loss. Both the student and discriminator losses are derived from the same adversarial objective function, fostering a min-max relationship in their training, as shown in Eq~\ref{minmax}.

\begin{equation}
\begin{split}
      minmax \quad (G, F) = \mathbb{E}_{x,y \sim p_{data}(x,y)}[\log F(x, y)] +\\
      \mathbb{E}_{x \sim p_{data}(x)} [\log(1 - F(x, G(x)))]
\end{split}
\label{minmax}
\end{equation}

Eq~\ref{minmax} represents the adversarial objective function in a game theory setting, where \(G\) is the student model, \(F\) is the discriminator, \(x\) is the input image, and \(y\) is the ground truth. The loss is designed to train both the student and the discriminator simultaneously. The overall discriminator loss $L_{disc}$ and adversarial loss of the student model $L_{adv}$ encapsulate the adversarial interplay within our segmentation framework for three inputs see in Eq ~\ref{disc}.

\begin{equation}
L_{disc} \sim L_{adv} = (L_{dat1} + L_{dat2}) / 2
\label{disc}
\end{equation}

We utilize binary cross-entropy loss for each dataset input, denoted as \(L_{dat}\). In \(L_{disc}\), the adversarial loss of (1), representing the labeled dataset features output from the student model, assesses the authenticity of the inputs labeled with 1's. However, the adversarial loss for (2) is computed with 0's, indicating the unlabeled dataset features. Similarly, we compute \(L_{adv}\) of the student model using an approach where both the adversarial losses of (1) and (2) are evaluated with 1's, indicating the extraction of the same feature embeddings for both morphologically similar datasets. Consequently, during training, we encourage the discriminator to distinguish between the different outputs as effectively as possible, while simultaneously motivating the student model to produce outputs that closely resemble reality.

\section{Experiments}
\subsection{Implementation Details}

In our architecture, our student model's two decoders for regularization enhancements, feature dropouts and feature noise, are applied to each feature layer before the decoder. The exponential moving average (EMA) decay rate ($\alpha$) undergoes temporal changes based on the function $\alpha = \min(1 - 1 / (\text{epoch} + 1), 0.95)$. For the teacher model, four dropout layers with a rate of 0.5 are introduced at the last U-Net decoder layer, enabling it to serve as an MC dropout for the vessel-unveiling module. Moreover, the discriminator in our setup comprises three convolutional layers with increasing filters, each followed by leaky ReLU activation. Gradual increases in the number of filters help capture complex patterns and features. Batch normalization after each convolutional layer ensures stable training and improved generalization. The output layer of the discriminator employs a sigmoid activation function.

The experiments are conducted on a server equipped with NVIDIA GeForce GTX 1080 Ti GPUs, using the PyTorch framework. The initial learning rate is set to 1e-4 for both models, and the Adam algorithm with a momentum of 0.9 is used to optimize both the student segmentation and the discriminator models. We perform K = 8 times stochastic forward passes in MC dropout. The model was trained for 250 epochs. During training, the batch size for labeled and unlabeled input was set to 10. To address the high time complexity associated with the original image dimensions during training, we adopted a patching strategy, breaking images into 400x400 sections. 

We utilize a random mix of four data augmentation techniques: random crop, horizontal and vertical flip, and random rotation, which are applied to both labeled and unlabeled data. Moreover, for unlabeled images, we introduce soft augmentation through color jitter and random grayscale. 

\subsection{Dataset}

In our model evaluation, we conduct a systematic assessment across two diverse publicly available medical image segmentation datasets: CHASEDB~\cite{b27}, STARE~\cite{b28}, and our in-house ROP dataset, to compare the quantitative segmentation results. These datasets present a wide spectrum of medical imaging scenarios, encompassing various segmentation challenges. The CHASEDB dataset consists of 28 healthy retinal images, capturing the eyes of 14 multi-ethnic children. STARE comprises 20 retinal images, including 10 healthy and 10 heterogeneous pathology eyes. Both the CHASEDB and STARE datasets are manually annotated in both training and testing sets. Our ROP dataset features a collection of fundus images from preterm infants obtained at Taipei Chang Gung Memorial Hospital (CGMH), spanning the years 2018 to 2021. Note that our ROP dataset lacks pixel-wise annotations, yet these images include ROP stage information at the image level per patient. We use 90 fundus images as our unlabeled dataset for our semi-supervised segmentation. 
To fairly test our model on public and ROP images, we collaborated with doctors from CGMH to extract 12 vessel annotations from ROP images. These 12 manually annotated images will be used only for quantitative evaluation. For our publicly available datasets, we selected 5 images for each dataset to test our model.

\subsection{Evaluation Metrics}

To quantitatively evaluate the algorithms, we employed five evaluation indicators: Mean Intersection over Union (mIoU), Dice Score (DSC), Accuracy (Acc), Variation of Information (VOI) \cite{voi}, and Adjusted Rand Index (ARI) \cite{ari}. mIoU is calculated as the ratio of the intersection area between the predicted and ground truth masks to their union area. DSC measures the pixel-wise similarity between the predicted and ground truth masks. Acc measures the proportion of correctly classified pixels. VOI quantifies the average information gain between the predicted and ground truth segmentations. ARI assesses the similarity between two segmentations, taking into account possible permutations of class labels. 

\subsection{Baselines}

\begin{table*}[htbp]
    \caption{Quantitative comparison on ROP \& CHASEDB test dataset (\%). We use 18 images of the CHASEDB as labeled samples and 90 ROP images as unlabeled samples. Upperbound represents for fully-supervised training setting with only 18 CHASEDB images.}
    \centering
    \setlength{\tabcolsep}{5pt}
    \begin{tabular}{@{}l*{5}{c}@{\hspace{0.5cm}}*{5}{c}@{}} 
        \toprule
        \textbf{Model} & \multicolumn{5}{c}{\textbf{ROP Test Set}} & \multicolumn{5}{c}{\textbf{CHASEDB Test Set}} \\
        \cmidrule(lr){2-6} \cmidrule(lr){7-11}
        & \textbf{mIoU ($\uparrow$)} & \textbf{DSC ($\uparrow$)} & \textbf{Acc ($\uparrow$)} & \textbf{VOI ($\downarrow$)} & \textbf{ARI ($\uparrow$)} 
        & \textbf{mIoU ($\uparrow$)} & \textbf{DSC ($\uparrow$)} & \textbf{Acc ($\uparrow$)} & \textbf{VOI ($\downarrow$)} & \textbf{ARI ($\uparrow$)} \\
        \midrule
        UpperBound& 17.49  & 27.68 & 97.62 & 20.09 & 26.25 & 72.25 & 83.87 & 97.64 & 27.03 & 80.36 \\
        MT & 20.98 & 33.18 & 97.30 & 23.38 & 31.16 & 70.19 & 82.46 & 97.47 & 28.35 & 78.74 \\
        UAMT & 20.59 & 33.04 & 97.11 & 24.82 & 30.83 & 70.98 & 83.00 & 97.54 & 27.73 & 79.38 \\
        CPS & 24.80 & 38.52 & \underline{97.49} & \underline{22.67} & 36.43 & 71.14 & 83.10 & 97.51 & 28.29 & 79.42 \\
        CCT& 21.81 & 34.78 & 96.83 & 26.62 & 32.30 & 71.10 & 83.08 & 97.49 & 28.58 & 79.35 \\
        DCT& 24.14 & 38.07 & 97.23 & 24.62 & 35.72 & 70.53 & 82.70 & 97.52 & 27.86 & 79.05 \\
        UCC& \underline{24.90} &\underline{39.00} &\textbf{ 97.60} & \textbf{21.88} & \underline{36.97} & 68.83 & 81.51 & 97.38 & 28.75 & 77.71 \\
        \midrule
        Our Model& \textbf{29.59} & \textbf{45.24} & 97.06 & 26.78 & \textbf{42.44} & \textbf{72.65} & \textbf{84.14} & \textbf{97.63} & \textbf{27.49} & \textbf{80.59} \\
        \bottomrule
    \end{tabular}
    \label{rop_quant}
\end{table*}

\begin{figure*}[!hbt]
\centering
\centerline{\includegraphics[width=0.85\textwidth]{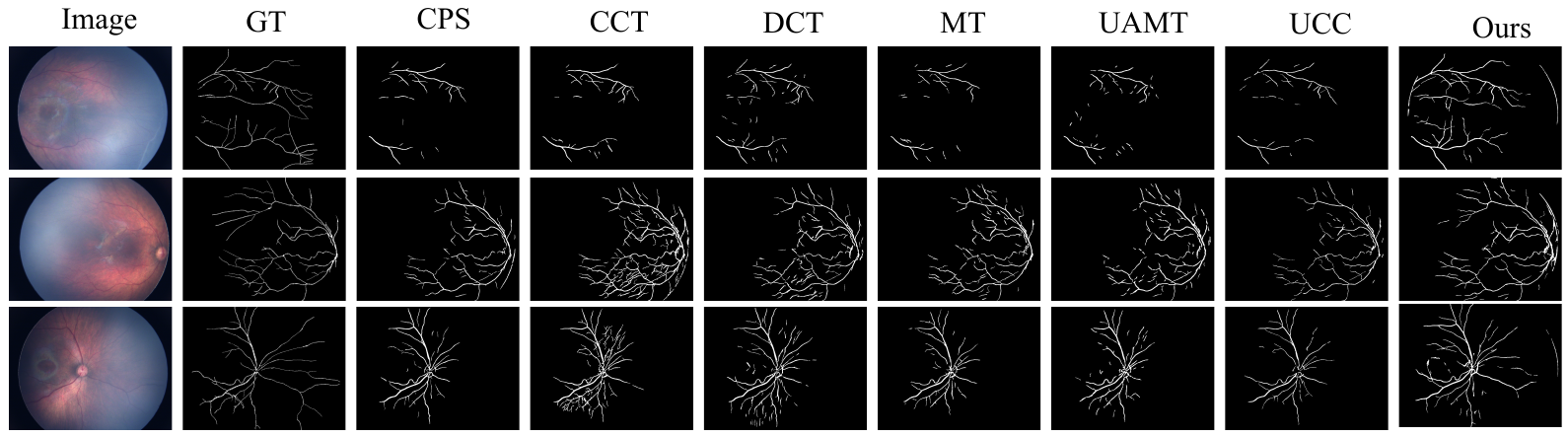}}
\caption{Visualization result of all baselines and our model on ROP test set with 18 CHASEDB labeled data and 90 ROP unlabeled images during training. The ROP vessels by their underdeveloped structure and background tissue colors, are harder to recognize than public datasets. Our uncertainty-aware vessel-unveiling module and latent feature alignments help to extend the vessel extraction through the end of the retina.}
\label{rop_viz}
\end{figure*}

To validate the feasibility of our model, we conducted a comprehensive comparison with six contemporary semi-supervised approaches, each representing a distinct paradigm in the realm of vessel segmentation. These methods include 1) Mean Teacher (MT)~\cite{b22}, adopting a teacher-student network to enforce consistency between perturbed inputs and the teacher network; 2) Uncertainty-Aware Mean Teacher (UAMT)~\cite{b11}, leveraging uncertainty maps derived from multiple forward passes to guide consistency training; 3) Cross-Pseudo Supervision (CPS)~\cite{b31}, a co-training method utilizing two parallel networks with one supervising the other based on pseudo-labels; 4) Cross-Consistency Training (CCT)~\cite{b58}, utilizing a shared encoder network with multiple decoders for consistency training, implemented with three decoders in our experiment; 5) Deep Co-Teaching (DCT)~\cite{b30}, employing the same network for ensuring consistency across different views of a given sample; and 6) Uncertainty-guided Crosshead Co-training (UCC)~\cite{b60}, employing two decoders with a shared encoder and robust data augmentation strategies to enforce consistency. In addition to these methods, we also evaluated the upper-bound setting where we trained a model using both labeled $(X_l)$ and unlabeled images $(X_{ul})$ in a fully supervised manner, including ground truth annotations for publicly available datasets.

\subsection{Results}

In Table \ref{rop_quant}, we present the quantitative results of our model, trained on both the CHASEDB and ROP datasets. This experiment involved using 18 labeled images from CHASEDB and 90 unlabeled images from the ROP dataset. To demonstrate the robustness and adaptability of our model in addressing both external and internal challenges, we compare its performance on both CHASEDB and ROP test sets. Remarkably, our model consistently outperforms other baselines across nearly all metrics, highlighting its ability to handle domain shifts and image variability effectively. Despite incorporating 90 unlabeled images from a different domain, our model maintains stable results on the CHASEDB test set compared to other baselines. Notably, in the ROP test set, our model achieved a Dice score of 45.24\%, outperforming the second-best result by more than 6\%. This success underscores the model's capacity to extract valuable information from unlabeled data by leveraging the morphological similarities across different fundus images.

In Figure~\ref{rop_viz}, we visually compare the vessel structures from three ROP test images using our proposed model trained on 18 labeled CHASEDB and 90 unlabeled ROP images with other baselines. Our vessel-unveiling method significantly improves vessel extraction in ROP images, particularly in extracting longer vessel segments. Even though the labeled dataset includes only OD-centered images, our model effectively extracts vessel structures in different fields of view, demonstrating the benefit of feature alignment across domains.

We further demonstrate the success of our semi-supervised modeling by presenting results for CHASEDB and STARE datasets in Tables \ref{chasedb} and \ref{stare}, respectively. We performed two experiments on each dataset with varying numbers of labeled and unlabeled input images to examine the label-unlabeled trade-off.

Quantitative performance comparisons on the CHASEDB dataset, presented in Table \ref{chasedb}, show that our model consistently achieves competitive results across all scenarios involving labeled and unlabeled sets. Our model outperforms baseline methods in terms of metrics such as mIoU, DSC, Accuracy, VOI, and ARI, achieving the best mIoU scores of 70.99\% and 72.05\% in both experimental settings. Table \ref{stare} quantitatively compares the results obtained on the STARE test dataset. Here, our model again outperforms baseline methods, achieving mIoU scores of 70.13\% and 71.41\% in two experimental settings, demonstrating consistent superiority across different labeled-unlabeled configurations. The results from both datasets indicate that our model exhibits strong robustness across various labeled-unlabeled sample distributions and maintains superior segmentation performance. 

\begin{table*}[!htbp]
    \caption{Quantitative Comparison of Experiments on CHASEDB Test Dataset With Two Different Labeled-Unlabeled Images (\%). The best values are in bold, and the second-best values are underlined. Upper Bound is fully supervised with 18 training images. }
    \centering
    \setlength{\tabcolsep}{5pt}
    \begin{tabular}{@{}l*{5}{c}@{\hspace{0.5cm}}*{5}{c}@{}} 
        \toprule
        \textbf{Model} & \multicolumn{5}{c}{\textbf{3 labeled, 15 unlabeled samples}} & \multicolumn{5}{c}{\textbf{5 labeled, 13 unlabeled samples}} \\
        \cmidrule(lr){2-6} \cmidrule(lr){7-11}
        & \textbf{mIoU ($\uparrow$)} & \textbf{DSC ($\uparrow$)} & \textbf{Acc ($\uparrow$)} & \textbf{VOI ($\downarrow$)} & \textbf{ARI ($\uparrow$)} 
        & \textbf{mIoU ($\uparrow$)} & \textbf{DSC ($\uparrow$)} & \textbf{Acc ($\uparrow$)} & \textbf{VOI ($\downarrow$)} & \textbf{ARI ($\uparrow$)} \\
        \midrule
        UpperBound & 72.25 & 83.87 & 97.64 & 27.03 & 80.36 & 72.25 & 83.87 & 97.64 & 27.03 & 80.36 \\
        MT         & 70.19 & 82.47 & 97.28 & 30.72 & 78.43 & 70.81 & 82.89 & 97.38 & 29.80 & 78.99 \\
        UAMT       & 69.77 & 82.18 & 97.18 & 31.45 & 78.00 & \underline{71.61} & \underline{83.44} & \underline{97.42} & 29.57 & \underline{79.58} \\
        CPS        & 70.51 & 82.69 & \underline{97.34} & \underline{30.17} & 78.74 & 71.14 & 83.11 & 97.41 & \underline{29.52} & 79.26 \\
        CCT        & 70.04 & 82.36 & 97.22 & 31.16 & 78.24 & 70.14 & 82.43 & 97.28 & 30.63 & 78.40 \\
        DCT        & \underline{70.69} & \underline{82.81} & 97.32 & 30.37 & \underline{78.81} & 70.97 & 83.00 & 97.39 & 29.67 & 79.13 \\
        UCC        & 67.14 & 80.33 & 97.06 & 32.25 & 76.07 & 69.06 & 81.69 & 97.12 & 32.01 & 77.43 \\
        \midrule
        Our Model  & \textbf{70.99} & \textbf{83.02} & \textbf{97.37} & \textbf{30.00} & \textbf{79.09} & \textbf{72.05} & \textbf{83.74} & \textbf{97.50} & \textbf{28.82} & \textbf{80.00} \\
        \bottomrule
    \end{tabular}
    \label{chasedb}
\end{table*}

\begin{figure*}[!hbt]
\centering
\centerline{\includegraphics[width=0.85\textwidth]{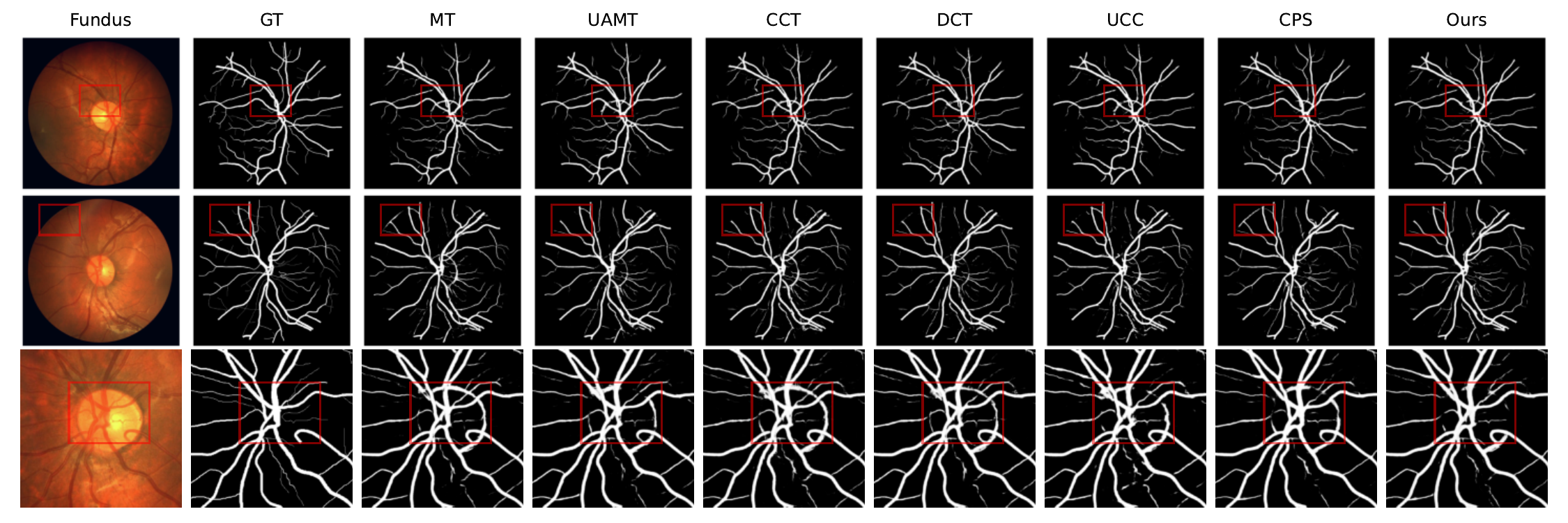}}
\caption{Visualization result of different methods on CHASEDB test set with 5 labeled data during training. To illustrate the discrepancy between other baselines, we set bounding boxes in the color red to share false positive regions.}
\label{chase_viz}
\end{figure*}

Additionally, Figure~\ref{chase_viz} shows the results from the CHASEDB dataset, where our model was trained on 5 labeled and 13 unlabeled CHASEDB images. Our model successfully induces better vessel structures, with increased length, reduced instances of vessel breakage, and more accurate segmentation around the central disk area compared to baseline methods. These results further validate the effectiveness of our framework in improving vessel segmentation.

\begin{table*}[!htbp]
    \caption{Quantitative Comparison of Experiments on STARE Test Dataset With Two Different Labeled-Unlabeled Images (\%). The best values are in bold and the second-best values are underlined. Upper Bound is fully supervised with 12 training images. }
    \centering
    \setlength{\tabcolsep}{5pt}
    \begin{tabular}{@{}l*{5}{c}@{\hspace{0.5cm}}*{5}{c}@{}} 
        \toprule
        \textbf{Model} & \multicolumn{5}{c}{\textbf{2 labeled, 10 unlabeled samples}} & \multicolumn{5}{c}{\textbf{4 labeled, 8 unlabeled samples}} \\
        \cmidrule(lr){2-6} \cmidrule(lr){7-11}
        & \textbf{mIoU ($\uparrow$)} & \textbf{DSC ($\uparrow$)} & \textbf{Acc ($\uparrow$)} & \textbf{VOI ($\downarrow$)} & \textbf{ARI ($\uparrow$)} 
        & \textbf{mIoU ($\uparrow$)} & \textbf{DSC ($\uparrow$)} & \textbf{Acc ($\uparrow$)} & \textbf{VOI ($\downarrow$)} & \textbf{ARI ($\uparrow$)} \\
        \midrule
        UpperBound & 72.12 & 83.74 & 97.56 & 27.95 & 80.08 & 72.12 & 83.74 & 97.56 & 27.95 & 80.08 \\
        MT         & 67.85 & 80.83 & 96.81 & 34.35 & 76.11 & 68.70 & 81.42 & 96.92 & 33.45 & 76.84 \\
        UAMT       & \underline{68.30} & \underline{81.16} & 96.76 & 34.43 & \underline{76.35} & 69.16 & 81.74 & 97.01 & 32.66 & 77.29 \\
        CPS        & 67.19 & 80.37 & 96.61 & 35.46 & 75.36 & \underline{70.90} & \underline{82.96} & \underline{97.18} & \underline{31.21} & \underline{78.72} \\
        CCT        & 68.24 & 81.11 & 96.71 & 34.63 & 76.22 & 67.03 & 80.24 & 96.63 & 35.45 & 75.28 \\
        DCT        & 67.51 & 80.59 & \underline{96.87} & \underline{33.91} & 75.98 & 69.89 & 82.26 & 97.10 & 31.97 & 77.93 \\
        UCC        & 64.78 & 78.61 & 96.34 & 37.79 & 73.30 & 68.76 & 81.46 & 97.02 & 32.80 & 77.05 \\
        \midrule
        Our Model  & \textbf{70.13} & \textbf{82.40} & \textbf{97.29} & \textbf{30.48} & \textbf{78.36} & \textbf{71.41} & \textbf{83.30} & \textbf{97.33} & \textbf{30.07} & \textbf{79.28} \\
        \bottomrule
    \end{tabular}
    \label{stare}
\end{table*}

The experimental results across these datasets show that our model excels in utilizing labeled and unlabeled data more effectively compared to other baselines. Our approach consistently achieves results that are close to the upper bound, reflecting the model's efficiency in leveraging available data. Despite the challenge of working with limited labeled samples, our model demonstrates adaptability and superior performance, showcasing its robustness and efficiency in data utilization. Interestingly, while our model consistently achieves the best performance, the second-best results vary among baseline models, indicating that no single baseline consistently performs well across all metrics, unlike our approach.

\subsection{Ablation studies} \label{ablation_study}

Using the CHASEDB dataset, which consists of 5 labeled and 13 unlabeled images, we systematically evaluate the incremental contributions of our key modules: 1) A student model equipped with three decoders, 2) Integration of the Teacher model with the adversarial architecture, 3) A discriminator for feature alignment, and 4) A vessel-unveiling module. We progressively added or replaced each proposed module within an existing standard framework to assess their individual and collective impacts, as illustrated in Table \ref{module}.

Model I serves as the baseline, representing a basic adversarial learning architecture with a standard U-Net as the student model and a discriminator with a real/fake output. Transitioning to Model II, an enhanced variant incorporating three decoders into the student model, we observe a notable increase across all evaluation metrics. The introduction of the teacher model in Model III, while retaining the student and discriminator from Model II, enhances several, but not all, performance metrics, including mIoU, DSC, and Acc. In Model IV, we adjusted the adversarial learning framework to focus the discriminator on feature alignment rather than the usual real/fake discrimination. This adjustment emphasizes the importance of the discriminator in feature alignment, resulting in improved performance across all metrics. Finally, Model V, augmented with an vessel-unveiling module in addition to the components of Model IV, achieves the highest segmentation scores across all metrics, clearly demonstrating the performance boost conferred by enhancing the teacher model with the vessel-unveiling module.

\begin{table}[!hbt]
    \caption{Analyzing Module Effects within Our Model on ChaseDB Dataset (\%).}
    \centering
    \setlength{\tabcolsep}{5pt} 
    \begin{tabular}{@{}l*{5}{c}@{}}
        \toprule
        \textbf{Model} & \textbf{mIoU ($\uparrow$)} & \textbf{DSC ($\uparrow$)} & \textbf{Acc ($\uparrow$)} & \textbf{VOI ($\downarrow$)} & \textbf{ARI ($\uparrow$)} \\
        \midrule
        model I & 69.67 & 82.10 & 97.19 & 31.41 & 77.94 \\
        model II & 70.53 & 82.69 & 97.38 & 29.70 & 78.82 \\
        model III & 70.58 & 82.74 & 97.35 & 30.03 & 78.81 \\
        model IV & 71.28 & 83.22 & 97.42 & 29.42 & 79.38 \\
        model V & \textbf{72.05} & \textbf{83.74} & \textbf{97.50} & \textbf{28.82} & \textbf{80.00} \\
        \bottomrule
    \end{tabular}
    \label{module}
\end{table}

\section{Downstream Task - Image Classification}

We extend our innovative semi-supervised vessel segmentation model to a multi-stage classification task on our ROP dataset, aiming to use the vessels extracted from preterm fundus images for diagnosing ROP disease. Our objective is to evaluate how these vessel structures, extracted without explicit annotations and from a domain different from the labeled datasets, can enhance ROP diagnosis for early intervention, thus potentially preserving infant vision.

Traditional ROP diagnosis relies on identifying stages based on features such as ridges and demarcations, with vessel abnormalities playing a crucial role in determining later stages. Our study seeks to leverage vessel structure information for improved ROP multi-stage classification, potentially discovering new biomarkers. To our knowledge, no prior studies have explored multi-class classification with or without vessel masks in pediatric retinopathy, making our approach both novel and significant.

\subsection{Dataset}

For this downstream task, we utilize a dataset from the same hospital (CGMH) as our in-house vessel segmentation dataset, but from a different cohort of patients, spanning the years 2020 to 2022. The dataset includes 3,873 fundus images taken from preterm infants, along with associated ROP stage information. It covers 217 individual patients, with an average of approximately 5 images per eye. The dataset is annotated with four stages of ROP: no ROP, stage 1, stage 2, and stage 3. The image distribution includes 1,548 images with no ROP, 915 with Stage 1 ROP, 897 with Stage 2 ROP, and 513 with Stage 3 ROP.

\subsection{Classification Model}

We employ a pre-trained EfficientNet~\cite{b61} model for image classification, applied independently to both the original fundus images and the vessel masks generated by our segmentation model. Additionally, we introduce a fusion model that processes both the fundus image and the vessel mask simultaneously. The fusion model leverages EfficientNet to extract features from both the fundus images and vessel masks, which are then combined with a controlled level of contamination. Two fully connected layers are added after the fusion step to refine the combined features and produce the final classification results.

We first use our segmentation model (trained on 18 labeled CHASEDB images and 90 unlabeled ROP images) to generate vessel masks for all 3,873 fundus images. We then perform three training experiments: one with the original fundus images, one with our vessel masks, and one using the fusion model.

\subsection{Results and Discussion}

Table \ref{tab:ai-rop-results} presents the results, highlighting the superior performance of the fusion model, which outperforms the fundus image-only model by 2\% in both Accuracy and F1-Score and by 4\% in Precision. These results demonstrate the added value of vessel mask extraction in enhancing ROP detection.


\begin{table}[!hbt]
    \caption{Multiclass Classification for different inputs.}
    \centering
    \setlength{\tabcolsep}{5pt}
    \begin{tabular}{@{}l*{5}{c}@{}}
        \toprule
        \textbf{Inputs} & \textbf{Accuracy} & \textbf{Precision} & \textbf{Recall} & \textbf{F1-Score} \\
        \midrule
        Fundus Images &  0.7476 & 0.7161 & \textbf{0.7188} & 0.7174 \\
        vessel\_mask (Ours) & 0.6388 & 0.6154 & 0.5914 & 0.6031 \\
        \midrule
        Fundus \& vessel\_mask & \textbf{0.7638} & \textbf{0.7579} & \underline{0.7144} & \textbf{0.7355} \\
        \bottomrule
    \end{tabular}
    \label{tab:ai-rop-results}
\end{table}

Our findings underscore the potential of using vessel extraction, including choroidal vessels, in improving ROP stage diagnosis. This opens avenues for further exploration in clinical diagnostic methodologies, particularly in early-stage interventions for preserving vision in preterm infants.

\section{Conclusion}

In this work, we propose a novel semi-supervised vessel segmentation model to advance the study of ROP without the labor-intensive process of manual vessel annotation. Our model bridges the significant gap between public labeled datasets and unlabeled ROP images by combining two powerful strategies: a vessel-veiling module in the teacher model and unified domain adversarial learning, forming a robust semi-supervised segmentation framework. The vessel-unveiling module reveals the obscured and hard-to-see vessel structures and also enhances the reliability of predictions in challenging areas. The adversarial training aligns feature representations between labeled and unlabeled datasets, despite variations in domains, age groups, and diseases, creating more robust and generalizable vessel segmentations.

We validated our model on multiple public and in-house ROP datasets, demonstrating its effectiveness across a range of evaluation metrics. The model's ability to leverage unlabeled data and adapt to different domains underscores its robustness, as shown by consistently superior performance over baseline methods. Furthermore, we extended the model to a downstream task of ROP multi-stage classification using vessel masks extracted by our segmentation model. The classification results demonstrated the utility of our model in enhancing diagnostic accuracy, particularly when combining vessel information with fundus images. 
Overall, our work not only addresses the challenge of limited labeled data for ROP vessel segmentation but also presents a viable solution for improving diagnostic processes through semi-supervised learning, paving the way for further exploration in pediatric ophthalmology.

\section{ACKNOWLEDGMENT}

We extend our deepest gratitude to Dr. Alex Po-Yi Wu and Dr. Dannie Chung-Ting Wang for their invaluable contributions to the evaluation process of the ROP vessel extraction phase. 
This work was partly supported by the grants PSC-CUNY Research Award 65406-00 53, NSF CCF-2144901, and NIGMS R01GM148970.




\end{document}